\documentclass[11pt,a4paper]{article}
\usepackage[hyperref]{emnlp2018}
\usepackage{times}
\usepackage{latexsym}
\usepackage{times}
\usepackage{latexsym}
\usepackage{amsmath}
\usepackage{booktabs}
\usepackage{url}
\usepackage{pgfplots}
\usepackage{pgfplotstable}
\usepackage{booktabs}
\usepackage{array}
\usepackage{colortbl}
\usepackage{filecontents}
\usetikzlibrary{patterns}
\usetikzlibrary{calc,positioning}
\usepgfplotslibrary{statistics}
\usepgfplotslibrary{groupplots}
\usepackage{tikzsymbols}
\usepackage{multicol}
\usepackage{ulem}
\usepackage{microtype}
\usepackage{paralist}
\usepackage{enumitem}
\setlist{noitemsep}
\usepackage{breqn}
\usepackage{amssymb}
\usepackage{subcaption}

\usepackage{url}
\usepackage{amsmath}
\usepackage{booktabs}

\newcommand{\src}[0]{{src}}
\newcommand{\pe}[0]{{pe}}
\newcommand{\mt}[0]{{mt}}

\aclfinalcopy 

\title{MS-UEdin Submission to the WMT2018 APE Shared Task: \\Dual-Source Transformer for Automatic Post-Editing}

\author{Marcin Junczys-Dowmunt \\
  Microsoft \\
  Redmond, WA 98052, USA \\
  {\tt marcinjd@microsoft.com} \\\And
  Roman Grundkiewicz \\
  University of Edinburgh \\
  10 Crichton St, Edinburgh EH8 9AB, Scotland \\
  {\tt rgrundki@inf.ed.ac.uk} \\}

\date{}

\begin{document}
\maketitle
\begin{abstract}

This paper describes the Microsoft and University of Edinburgh submission to the Automatic Post-editing shared task at WMT2018. Based on training data and systems from the WMT2017 shared task, we re-implement our own models from the last shared task and introduce improvements based on extensive parameter sharing. 
Next we experiment with our implementation of dual-source transformer models and data selection for the IT domain.
Our submissions decisively wins the SMT post-editing sub-task establishing the new state-of-the-art and is a very close second (or equal, 16.46 vs 16.50 TER) in the NMT sub-task.
Based on the rather weak results in the NMT sub-task, we hypothesize that neural-on-neural APE might not be actually useful.
\end{abstract}

\section{Introduction}

This paper describes the Microsoft (MS) and University of Edinburgh (UEdin) submission to the Automatic Post-editing shared task at WMT2018 \cite{chatterjee-EtAl:2018:WMTAPE}. Based on training data and systems from the WMT2017 shared task \cite{bojar-EtAl:2017:WMT1}, we re-implement our own models from the last shared task \cite{junczysdowmunt-grundkiewicz:2017:WMT,I17-1013} and introduce a few small improvements based on extensive parameter sharing. 
Next, we experiment with our implementation of dual-source transformer models which have been available in our NMT toolkit Marian \cite{marian} since version v1.0 (November 2017). We believe this is one of the first descriptions of such an architectures for Automatic Post-Editing (APE) purposes, but similar approaches have been used for two-step decoding, for instance in \newcite{DBLP:journals/corr/abs-1803-05567}.
We further extend this model to share parameters across encoders with improved results for APE. 

Our submissions decisively wins the SMT post-editing sub-task  establishing the new state-of-the-art and is a very close second (or equal, 16.46 vs 16.50 TER) in the NMT sub-task.\footnote{We did not make the models available, but researchers interested in reproducing these results are encouraged to contact one or both of the authors. We will be happy to help. The used architectures are available in Marian: \url{https://marian-nmt.github.io}}

\section{Training, development, and test data}

We perform all our experiments with the official WMT-2018 automatic post-editing data and the respective development and test sets. The training data consists of a small set of post-editing triplets $(\src,\mt,\pe)$, where $\src$ is the original English text, $\mt$ is the raw MT output generated by an English-to-German system, and $\pe$ is the human post-edited MT output. The MT system used to produce the raw MT output is unknown, as is the original training data. The task consists of automatically correcting the MT output so that it resembles human post-edited data. The main task metric is TER \cite{citeulike:1874459} --- the lower the better --- with BLEU \cite{Papineni:2002:BMA:1073083.1073135} as a secondary metric.

To overcome the problem of too little training data, \newcite{junczysdowmunt-grundkiewicz:2016:WMT} --- the authors of the best WMT-2016 APE shared task system --- generated large amounts of artificial data via round-trip translations.
The artificial data has been filtered to match the HTER statistics of the training and development data for the shared task and was made available for download. 

The organizers also made available a large new resource for APE training, the eSCAPE corpus \cite{DBLP:journals/corr/abs-1803-07274}, which contains triplets generated from SMT and NMT systems in separate data sets.


To produce our final training data set we oversample the original training data 20 times and add both artificial data sets. This results in a total of slightly more than 5M training triplets.
We validate on the development set for early stopping and report results on the WMT-2016 APE test set. The data is already tokenized. Additionally we truecase all files and apply segmentation into BPE subword units \cite{sennrich2016bpe}. We reuse the subword units distributed with the artificial data set.


\section{Experiments}

\begin{table*}[t]
\centering
\begin{tabular}{p{5.5cm}cccccc}
\toprule
& \multicolumn{2}{c}{dev 2016} & \multicolumn{2}{c}{test 2016} & \multicolumn{2}{c}{test 2017}\\
Model & \textbf{TER$\downarrow$} & BLEU$\uparrow$ & \textbf{TER$\downarrow$} & BLEU$\uparrow$ & \textbf{TER$\downarrow$} & BLEU$\uparrow$ \\
\midrule
Uncorrected & 24.81 & 62.92 & 24.76 & 62.11 & 24.48 & 62.49 \\

\midrule
WMT17: FBK Primary & 19.22 & 71.89 & 19.32 & 70.88 & 19.60 & 70.07 \\
WMT17: AMU Primary &   --- &   --- & 19.21 & 70.51 & 19.77 & 69.50 \\

\midrule
Baseline (single model) & 19.77 & 70.54 & 20.10 & 69.25 & 20.43	 & 68.48 \\
\quad +Tied embeddings & 19.39 & 70.70 & 19.82 & 68.87 & 20.09 & 69.06 \\
\quad +Shared encoder & 19.23 & 71.14 & 19.44 & 70.06 & 20.15 & 69.04 \\ \midrule
Transformer-base (Tied+Shared) & 18.73 & 71.71 & 18.92 & 70.86 & 19.49 & 69.72 \\ 
Transformer-base x4 & 18.22 & 72.34 & 18.86 & 71.04 & 19.03 & 70.46 \\

\bottomrule
\end{tabular}
\caption{Experiments with WMT 2017 data, correcting a phrase-base system.}\label{tab:results}
\end{table*}

During the WMT2017 APE shared task we submitted a dual-source model with soft and hard attention which placed second right after a very similar dual-source model by the FBK team. We include the performance of those models based on the shared task descriptions in Table \ref{tab:results}, systems WMT17:FBK and WMT17:AMU (ours).

We mostly worked on the APE sub-task for automatic post-editing for the SMT system. The system in the NMT sub-task seemed to have only small margins for improvements. 

\subsection{Baselines}

During the WMT2017 shared task on post-editing we made an error in judgment and submitted the weaker hard-attention model, in post-submission experiments we saw that a normal soft-attention model would have fared better. This was confirmed by the shared-task winner FBK and our own experiments. For this year, we first recreated our own dual-source model with soft attention (Baseline) and further experimented with parameter sharing:
\begin{itemize}
\item We first tie embeddings across all encoder instances, the decoder embedding layer and decoder output layer (transposed). This leads to visible improvements over our baseline across all test sets in terms of TER. 
\item Next, we share all parameters across encoders, despite the fact that these are encoding different language it seems that parameter sharing is generally beneficial. We see improvement across two test sets and roughly equal performance for the third. 
\end{itemize}

\begin{figure}[t]
\centering
\includegraphics[width=0.5\textwidth]{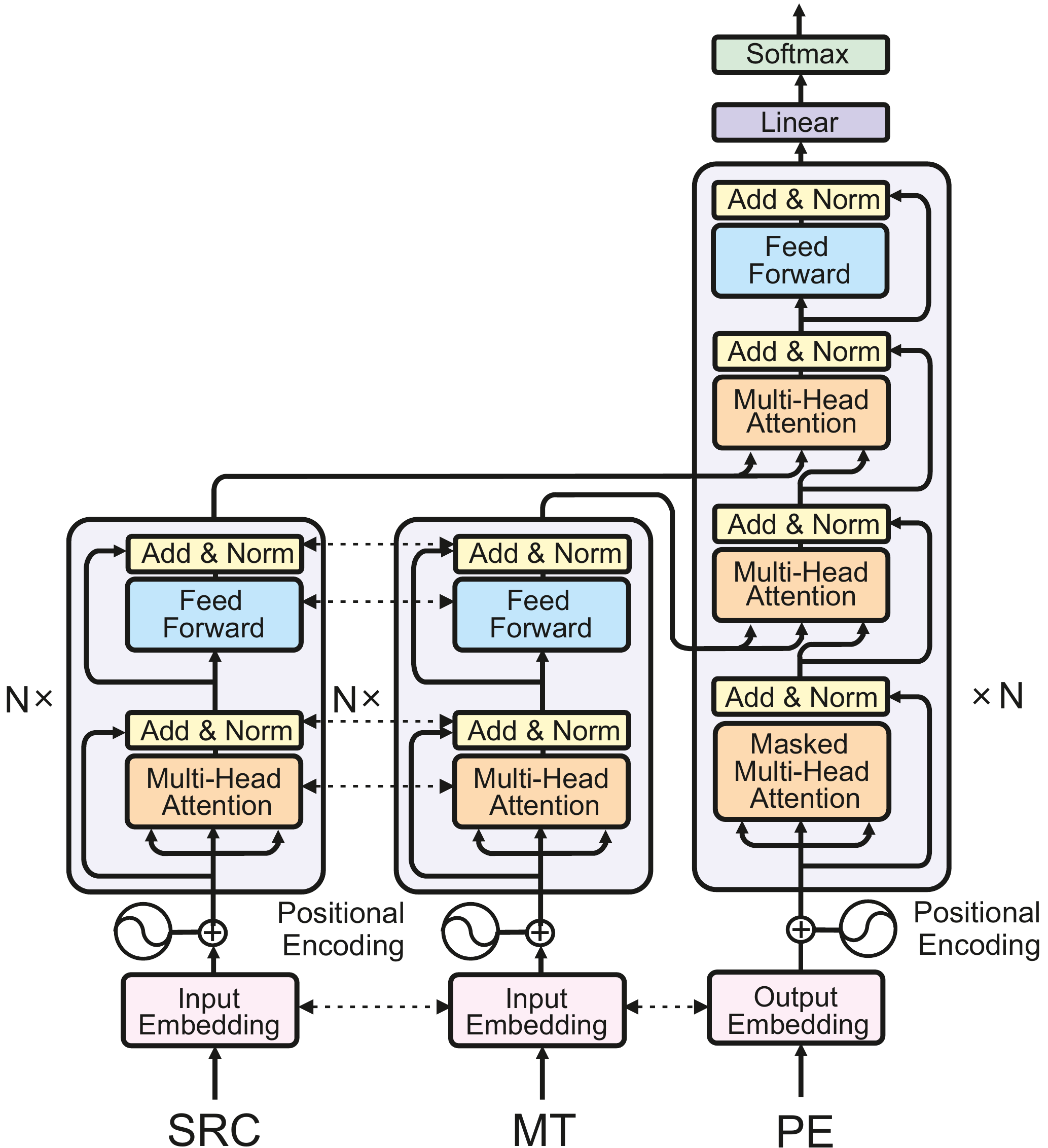}
\caption{Dual-source transformer architecture. Dashed arrows mark tied parameters between the two separate encoders and common embedding matrices for all encoders and the decoder.} \label{dual}
\end{figure}

\subsection{Dual-source transformer}

\begin{table*}[t]
\centering
\begin{tabular}{p{5.5cm}cccccc}
\toprule
& \multicolumn{2}{c}{dev 2016} & \multicolumn{2}{c}{test 2016} & \multicolumn{2}{c}{test 2017}\\
Model & \textbf{TER$\downarrow$} & BLEU$\uparrow$ & \textbf{TER$\downarrow$} & BLEU$\uparrow$ & \textbf{TER$\downarrow$} & BLEU$\uparrow$ \\
\midrule
Transformer all & 17.84 & 73.45 & 17.81 & 72.79 & 18.10 & 71.72 \\ 
Transformer 1M & 17.59 & 73.45 & 18.29 & 72.20 & 18.42 & 71.50 \\ 
Transformer 2M & 17.92 & 73.37 & 18.02 & 72.41 & 18.35 & 71.57\\ 
Transformer 4M & 17.75 & 73.51 & 17.89 & 72.70 & 18.09 & 71.78 \\ \midrule
\bf Transformer x4 (all above) & \bf 17.31 & \bf 74.14 & \bf 17.34 & \bf 73.43 & \bf 17.47 & \bf 72.84 \\ 

\bottomrule
\end{tabular}
\caption{Experiments with WMT 2017+eSCAPE data for SMT system.}\label{tab:results2}
\end{table*}

Figure~\ref{dual} illustrates the architecture of our dual-source transformer variant. We naturally extend the original architecture from \newcite{NIPS2017_7181} by adding another encoder and stacking an additional target-source multi-head attention component above the previous target-source multi-head attention component. This results in one target-source attention component per block for each encoder. As usual for the transformer architecture, each multi-head attention block is followed by a skip connection from the previous input and layer normalization. Each encoder corresponds exactly to the implementation from \newcite{NIPS2017_7181}, but with common parameters. Apart from these modifications, we follow the transformer-base configuration from \newcite{NIPS2017_7181}. This means that we tie source, target and output embeddings. 

We found earlier that sharing parameters between the encoders is beneficial for the APE task and apply the same modification to our architecture, marked by dashed arrows in Figure~\ref{dual}. The two encoders share all parameters, but still produce different activations and are combined in different places in the decoder. 

We briefly experimented with concatenating the encoder outputs instead of stacking (this would have been more similar to our work in \newcite{junczysdowmunt-grundkiewicz:2017:WMT,I17-1013}), but found this solution to underperform. We also replaced skip connections with gating mechanisms, but did not see any improvements.

The transformer architecture with its skip connections and normalization blocks can be seen to learn interpolation functions between layers that are not much different from gating mechanisms.  

A single model of this type outperforms already the complex APE ensembles from the previous shared task in terms of TER and is on par in terms of BLEU (Table \ref{tab:results}). An ensemble of four identical models trained with different random initializations strongly improves over last year's best models on all indicators. 

\subsection{Experiments with eSCAPE}

So far, we only trained on data that was available during WMT2017. This year, the task organizers added a new large corpus  created for automatic post-editing across many domains. We experimented with domain selection algorithms for this corpus and tried to find subsets that would be better suited to the given IT domain. We trained an 5-gram language model on a 10M words randomly sampled subset of the German IT training data and a similarly size language model on the eSCAPE data. Next we applied cross-entropy filtering \cite{moore-lewis:2010:Short} to produce domain scores. We sorted eSCAPE by these scores and selected different sizes of subsets. Smaller subsets should be more in-domain. We experimented with 1M, 2M, 4M and all sentences (nearly 8M). 
Results (Table~\ref{tab:results2}) remain however inconclusive. Adding eSCAPE to the training data was generally helpful, but we did not see a clear winner across subsets and test sets. In the end we use all the experimental models as components of a 4x ensemble. The different training sets might as well serve as additional randomization factors potentially beneficial for ensembling. 

\subsection{The NMT APE sub-task}
So far we reported only results for the SMT APE sub-task. For the NMT system we trained our transformer-base model on eSCAPE NMT data only. Including SMT-specific data seemed to be harmful. In the end we only applied an ensemble of 4 such models observing moderate improvements on the development data. It seemed that our system was quite good at correcting errors due to hallucinated BPE words. We believe that our shared embeddings/encoders were helpful here. This does however indicate that the corrected NMT system was not well designed as these errors could have been easily avoided by the original MT system.

Furthermore, our submission did only train for about one day, we would expect better results for a converged system, but we did not pursue this any further due to time constraints.

\section{Results and conclusions}

\begin{table}[t]\centering
\begin{subtable}[t]{7cm}\centering
\begin{tabular}{p{4cm}cc} \toprule
Systems & \textbf{TER$\downarrow$} & BLEU$\uparrow$ \\ \midrule
\bf MS-UEdin (Ours) & \bf 18.00 & \bf 72.52 \\
FBK  &18.62&	71.04\\
POSTECH &	19.63&	69.87\\
USAAR DFKI &	22.69&	66.16\\
DFKI-MLT &	24.19&	63.40\\
Baseline&	24.24	&62.99\\ \bottomrule
\end{tabular}
\caption{PBSMT sub-task}\label{pbmt}
\end{subtable}

\vspace{0.5cm}

\begin{subtable}[t]{7cm}\centering
\begin{tabular}{p{4cm}cc}\toprule
Systems & \textbf{TER$\downarrow$} & BLEU$\uparrow$ \\ \midrule
FBK & 16.46&	75.53 \\
\bf MS-UEdin (Ours)	& \bf 16.50& \bf	75.44 \\
POSTECH &16.70&	75.14 \\
Baseline&	16.84&	74.73 \\
USAAR DFKI&	17.23&	74.22 \\
DFKI-MLT&	18.84&	70.87 \\ \bottomrule
\end{tabular}
\caption{NMT sub-task}\label{nmt}
\end{subtable}
\caption{APE Results provided by shared task organizers. We only include best-scored results by each team, see \newcite{chatterjee-EtAl:2018:WMTAPE} for the full list of results.}
\label{org}
\end{table}

The organizers informed us about the results of our systems and we include the scores for the best system of each team in Table~\ref{org}. For full results with information concerning statistical significance see the full shared task description \cite{chatterjee-EtAl:2018:WMTAPE}. As expected, improvements are quite significant for the SMT-based system, and much smaller for the NMT-based system. Our submissions to the PBSMT sub-task strongly outperforms all submissions by other teams in terms of TER and BLEU and established the new state-of-the-art for the field. The improvements over the PBSMT baseline approach impressive 10 BLEU points.

For the NMT sub-task our submission places second with a 0.04 TER difference behind the leading submission. We would call this an equal result. This is interesting considering how little time and effort was spent on our NMT system compared to the SMT system. One day more or training time might have flipped these results. 

Based on the overall weak performance for the neural sub-task, we feel justified in not investing much time into that particular sub-task. We hypothesize that if the same amount of effort had been put into the NMT baseline as into the APE systems that were submitted to the task, none of the submissions (including our own) would have been able to beat that baseline. We saw obvious problems with BPE handling in the baseline which could have been easily fixed. It is probable that most of our improvements come from correcting those BPE errors.

We further believe that this might constitute the end of neural automatic post-editing for strong neural in-domain systems. The next shared task should concentrate on correcting general domain on-line systems. Another interesting path would be to make the original NMT training data available so that both, pure NMT systems and APE systems, can compete. This would show us where we actually stand in terms of feasibility of neural-on-neural automatic post-editing.

\section*{Acknowledgments}

We would like to thank Elena Voita and Rico Sennrich for sharing their transformer figures which we used as basis for our dual-source transformer illustration. We also thank Kenneth Heafield for his comments on the paper.

This work was partially funded by Facebook. The views and conclusions contained herein are those of the authors and should not be interpreted as necessarily representing the official policies or endorsements, either expressed or implied, of Facebook.

\bibliography{emnlp2018,ape}
\bibliographystyle{acl_natbib_nourl}

\end{document}